\PassOptionsToClass{nonacm}{acmart}
\documentclass[sigconf,screen]{acmart}
\usepackage{booktabs}
\usepackage{multirow}
\usepackage{pifont}
\usepackage{xspace}

\newcommand{\waf}{WAF\xspace}

\renewcommand\footnotetextcopyrightpermission[1]{}
\begin{document}

\title{Style over Substance: A Shortcut Audit of Emotion-Description Preference Evaluation}

\newcommand{\meraffiliation}{%
  \affiliation{%
    \institution{University of Chinese Academy of Sciences}
    \institution{Institute of Automation, Chinese Academy of Sciences}
    \city{Beijing}
    \country{China}}}

\author{Jiabing Yang}
\email{yangjiabing2025@ia.ac.cn}
\meraffiliation

\author{Yixiang Chen}
\email{yixiang.chen@cripac.ia.ac.cn}
\meraffiliation

\author{Yuan Xu}
\email{yuan.xu@nlpr.ia.ac.cn}
\meraffiliation

\author{Qisen Ma}
\email{maqisen2024@ia.ac.cn}
\meraffiliation

\author{Tao Yu}
\email{yutao2025@ia.ac.cn}
\meraffiliation

\author{Peiyan Li}
\email{peiyan.li@cripac.ia.ac.cn}
\meraffiliation

\author{Yingda Li}
\email{liyingda2025@ia.ac.cn}
\meraffiliation

\author{Yan Huang}
\authornotemark[2]
\makeatletter
\g@addto@macro\@authornotes{\footnotetext[2]{Corresponding author.}}
\makeatother
\email{yhuang@nlpr.ia.ac.cn}
\meraffiliation

\author{Liang Wang}
\email{wangliang@nlpr.ia.ac.cn}
\meraffiliation

\begin{abstract}
Preference over model-generated emotion descriptions is emerging as a standard evaluation metric for multimodal emotion understanding, exemplified by the MER2026 MER-Prefer track on EmoPrefer. Such benchmarks assume that predicting the preferred description requires grounded cross-modal understanding of the video. We conduct a systematic shortcut audit of EmoPrefer using content-blind probes, and demonstrate that a simple logistic regression utilizing only description length and generator identity---without ever processing the text, video, or audio---performs on par with LoRA-finetuned 7B text and audio-visual judges (65.8 vs.\ 66.8 \waf on EmoPrefer-V2). Analysis reveals that generator identity is 99.5\% recoverable from description text, every candidate pair contrasts two distinct generators, and the human preference labels align with a per-generator win-rate prior on 66\% of the evaluated pairs. Where the human label contradicts this prior, trained judges still side with the style prior 63--80\% of the time. On a length-matched subset that neutralizes verbosity bias, the tested media configurations yield no statistically significant improvement, and an ODIN-style diagnostic that decouples the style shortcut leaves its content head near chance. These results do not imply that human preferences are inherently stylistic or that the descriptions carry no emotional information; they show that the current scores can be reached without verifying either description against the video. We therefore recommend source-balanced pairing, strict length control, counter-stereotypical sliced reporting, and multi-annotator consensus for future cross-generator evaluations. Code is available at \url{https://github.com/jiabingyang01/EmoPrefer-Audit}.
\end{abstract}


\maketitle
\renewcommand{\shortauthors}{Yang et al.}
\pagestyle{plain}
\thispagestyle{plain}

\begin{figure*}[t]
  \centering
  \includegraphics[width=0.86\textwidth]{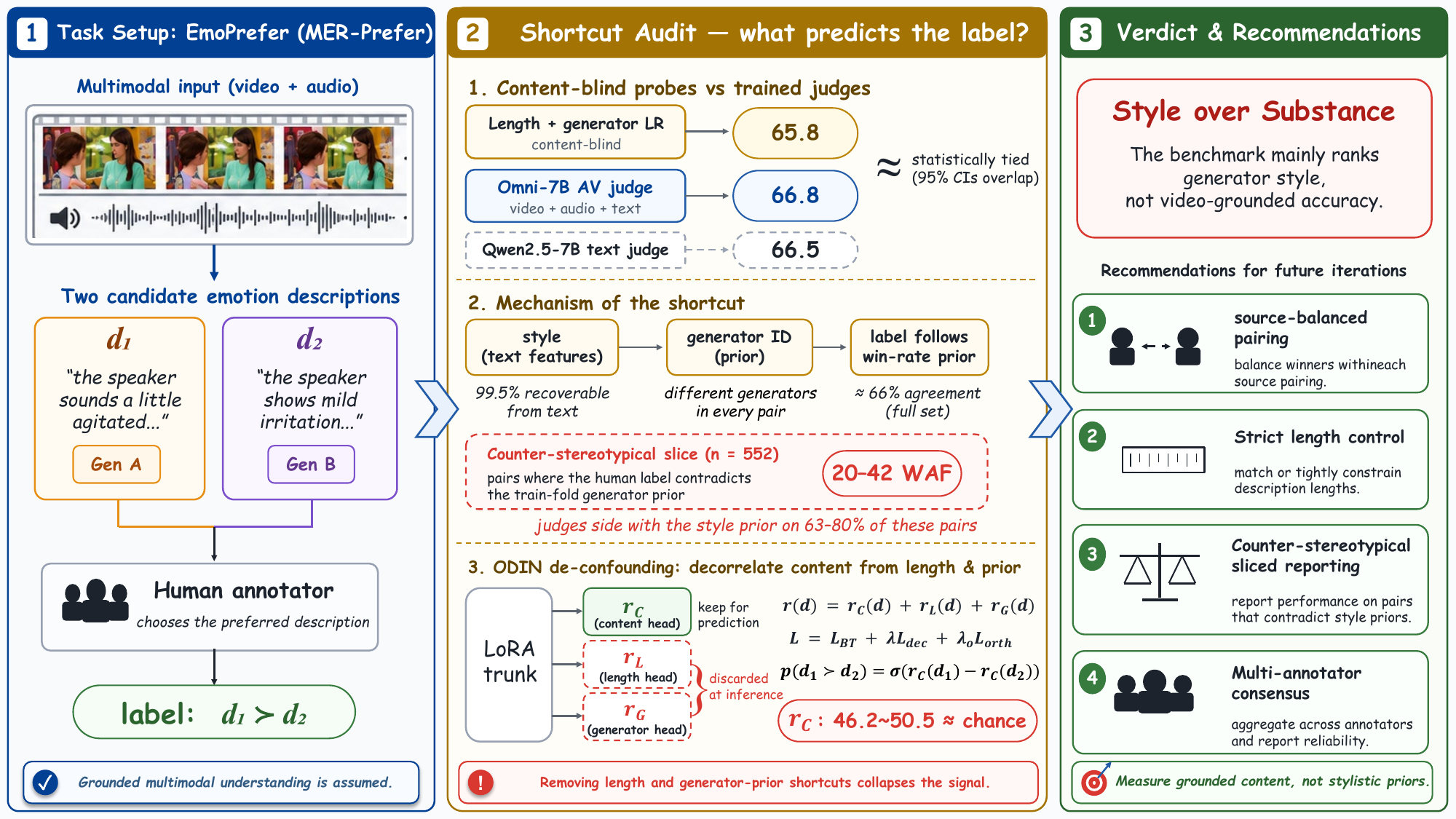}
  \Description{Three-stage overview: the preference task setup, the shortcut audit with content-blind probes and ODIN deconfounding, and the verdict with four dataset remedies.}
  \caption{Overview of our shortcut audit of EmoPrefer. \textbf{Stage 1 (task setup):} a judge receives the video, its audio, and two candidate emotion descriptions, and must predict the human-preferred one. \textbf{Stage 2 (audit):} content-blind probes match finetuned 7B judges; the shortcut works because writing style reveals the generator identity, and the labels follow a per-generator win-rate prior on 66\% of the evaluated pairs; judges collapse to 20--42 \waf where the label contradicts this prior, and ODIN-style deconfounding leaves the content head $r_C$ at chance. \textbf{Stage 3 (verdict):} the benchmark conflates generator-level preference with video-grounded accuracy, motivating four data-side fixes for future iterations.}
  \label{fig:pipeline}
\end{figure*}

\section{Introduction}
Open-vocabulary emotion understanding~\cite{lian2025affectgpt,lian2024mer} is increasingly evaluated via \emph{preference modeling}: given a video clip and two candidate emotion descriptions, human annotators select the more accurate description, and models are evaluated on their ability to predict this choice. The MER2026 challenge dedicates a specific track (MER-Prefer) to this task, utilizing the EmoPrefer datasets for training and held-out evaluation~\cite{lian2026mer}. The foundational assumption of such benchmarks is that accurately predicting human preference requires deep multimodal grounding---verifying the descriptions against facial expressions, vocal tones, and visual events in the video. The same preference signals increasingly supervise reward models for multimodal alignment~\cite{ouyang2022training,rafailov2023direct}, so what these labels actually measure matters beyond any single leaderboard or challenge.

This paper critically examines that assumption. Rather than optimizing for state-of-the-art scores, we investigate \emph{what information is actually necessary} to achieve the observed benchmark performance. We treat the benchmark itself as the object of study, auditing it through deliberately impoverished predictors~\cite{gururangan2018annotation,poliak2018hypothesis}, counter-stereotypical slicing~\cite{gardner2020evaluating}, and controlled debiasing interventions~\cite{chen2024odin} on finetuned 7B text and audio-visual judges. Figure~\ref{fig:pipeline} overviews this audit pipeline.

Our empirical audit reveals that the observed scores can be reached without verifying either description against the video. A lightweight logistic regression utilizing only 18 content-blind features (description lengths and one-hot generator identities) performs on par with finetuned Qwen2.5-7B text models~\cite{qwen2025qwen25technicalreport} and Qwen2.5-Omni audio-visual models~\cite{xu2025qwen25omni}, with overlapping 95\% confidence intervals. This holds true on both the full benchmark and a length-matched slice explicitly designed to remove verbosity shortcuts. We demonstrate that this occurs because generator identity strongly leaks into the description text (99.5\% recoverable via a simple bag-of-words classifier), and the benchmark's label distribution is heavily skewed toward a per-generator win-rate prior. Consequently, a predictor can reach the performance ceiling by implicitly identifying \emph{who wrote the description}.

We make three contributions. \ding{202}~We establish a \textbf{shortcut audit protocol} for pairwise preference benchmarks (content-blind probes, counter-stereotypical slices, and prior-following analysis) that is computationally lightweight and applicable to LLM-as-a-judge evaluations broadly. \ding{203}~We provide an \textbf{empirical diagnosis of EmoPrefer}, quantifying each shortcut with confidence intervals and showing through permutation and counterfactual interventions that the probes rest on generator identity rather than any hidden content correlate. \ding{204}~We present \textbf{controlled negative results} across trained and zero-shot judges from the Qwen and LLaVA families, showing that audio-visual grounding brings no detectable improvement on de-confounded slices and that zero-shot judges inherit the same prior without seeing a label.

\begin{figure*}[t]
  \centering
  \includegraphics[width=0.80\textwidth]{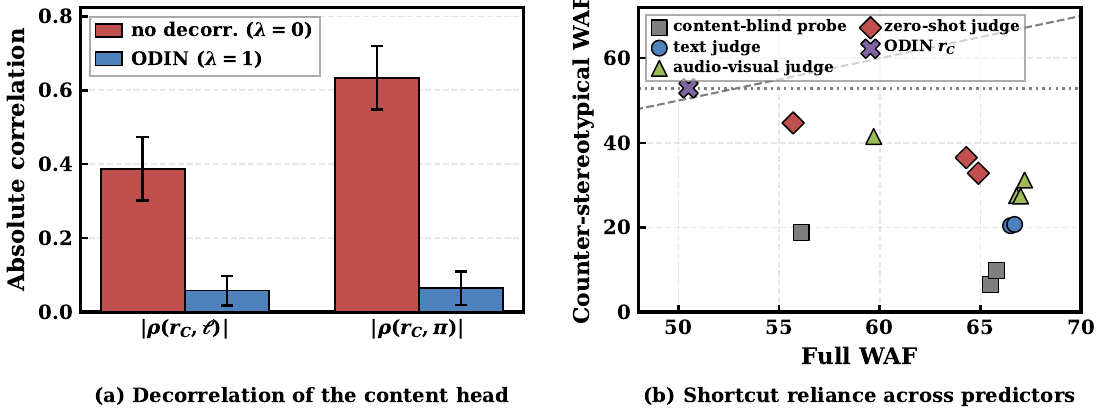}
  \Description{Two panels show ODIN head correlations before and after decorrelation, and full-set versus prior-conflicting performance for all predictors.}
  \caption{\textbf{(a)} Mean absolute head--confound correlations across folds; error bars show one standard deviation. Decorrelation suppresses length and prior correlations in $r_C$. \textbf{(b)} Full versus prior-conflicting \waf (dashed: equal performance; dotted: stratified-random reference). Most judges fall well below the diagonal; the deconfounded text head is near chance on both axes.}
  \label{fig:main}
\end{figure*}

\begin{table*}[t]
  \centering
  \caption{Two-class \waf [bootstrap 95\% CI] on EmoPrefer-V2 across different evaluation slices. Content-blind probes never see description text, video, or audio. The `Counter-stereotypical' slice consists of pairs where the human label contradicts the train-fold generator prior (a pure style prior scores 0 here by definition). The de-biased reward rows compare controls without decorrelation against ODIN content heads. Zero-shot judges span the Qwen and LLaVA architectures.}
  \label{tab:main}
    \begin{tabular}{llccc}
  \toprule
  & \textbf{Predictor} & \textbf{Full ($n{=}1625$)} & \textbf{Length-matched ($n{=}310$)} & \textbf{Counter-stereo.\ ($n{=}552$)} \\
  \midrule
  \multirow{4}{*}{\textbf{\shortstack[l]{Content-blind\\probes}}}
  & Length LR & 56.1 [53.3, 59.1] & 40.0 [33.7, 46.6] & 18.8 [15.3, 22.6] \\
  & Generator LR & 65.5 [63.1, 67.9] & 63.6 [58.2, 68.9] & 6.5 [4.5, 8.6] \\
  & Length{+}generator LR & 65.8 [63.2, 68.2] & 64.4 [58.9, 69.8] & 9.8 [7.4, 12.6] \\
  & Generator win-rate rule & 64.8 [62.3, 67.1] & 64.4 [59.2, 69.6] & 0.0 \\
  \midrule
  \multirow{2}{*}{\textbf{Text judges}}
  & Qwen2.5-7B (V2) & 66.5 [64.1, 68.9] & 65.6 [60.2, 71.0] & 20.4 [16.8, 23.8] \\
  & Qwen2.5-7B (V1{+}V2) & 66.7 [64.4, 69.1] & 65.2 [59.9, 70.7] & 20.7 [17.3, 24.2] \\
  \midrule
  \multirow{4}{*}{\textbf{\shortstack[l]{Audio-visual\\judges (Omni-7B)}}}
  & 12 frames {+} audio & 66.8 [64.4, 69.1] & 65.3 [60.0, 70.4] & 27.5 [23.7, 31.4] \\
  & 16 frames {+} audio & 67.2 [65.0, 69.6] & 64.9 [59.7, 70.2] & 31.2 [27.4, 35.3] \\
  & 8 frames, no audio & 67.0 [64.6, 69.2] & 65.0 [59.4, 70.3] & 27.4 [23.6, 31.1] \\
  & CoT (8frames, no audio) & 59.7 [57.3, 62.0] & 58.9 [53.3, 64.7] & 41.5 [37.3, 45.4] \\
  \midrule
  \multirow{4}{*}{\textbf{\shortstack[l]{De-biased\\reward}}}
  & Text, no decorrelation ($\lambda{=}0$) & 61.3 [58.9, 63.6] & 62.4 [56.9, 67.8] & 26.6 [22.9, 30.4] \\
  & Text, ODIN content head $r_C$~\cite{chen2024odin} & 50.5 [48.0, 52.8] & 50.5 [45.1, 56.3] & 52.9 [48.8, 57.1] \\
  & Omni, no decorrelation & 58.2 [55.8, 60.6] & 50.6 [44.7, 56.2] & 32.7 [28.7, 36.7] \\
  & Omni, ODIN content head $r_C$ & 46.2 [43.7, 48.5] & 45.8 [40.1, 51.8] & 58.4 [54.5, 62.3] \\
  \midrule
  \multirow{3}{*}{\textbf{\shortstack[l]{Zero-shot\\(cross-model)}}}
  & Omni-7B (official) & 64.3 [62.0, 66.6] & 64.8 [59.5, 70.3] & 36.5 [32.5, 40.8] \\
  & Qwen2.5-VL & 64.9 [62.6, 67.3] & 62.5 [56.8, 68.0] & 32.8 [28.7, 36.9] \\
  & LLaVA-NeXT-Video & 55.7 [53.3, 58.1] & 54.3 [48.6, 60.0] & 44.7 [40.4, 48.8] \\
  \bottomrule
  \end{tabular}
\end{table*}

\section{Related Work}
\paragraph{Shortcut learning and evaluator bias.}
Machine learning models tend to exploit whatever spurious feature most cheaply separates the training labels~\cite{geirhos2020shortcut,sagawadistributionally}. In NLP, hypothesis-only baselines famously exposed such annotation artifacts~\cite{gururangan2018annotation,poliak2018hypothesis,mccoy2019right}; our content-blind probes serve an analogous purpose, much like contrast sets and behavioral tests~\cite{gardner2020evaluating,ribeiro2020beyond}. Pairwise judging by humans and LLMs~\cite{zheng2023judging,liu2023g,kim2024prometheus} is notoriously susceptible to verbosity, position, and style biases~\cite{wang2024large,singhal2023long,liu2025rm,dubois2024length}. Recent efforts remove these confounds from the model side: ODIN disentangles a length head from a quality head~\cite{chen2024odin}; later work enforces counterfactual invariance~\cite{wang2025beyond,srivastavarobust} or debiases the training data~\cite{park2024offsetbias,liu2025rrm,ye2026rectifying}. These fixes assume the labels themselves are sound; in EmoPrefer the shortcut lives in the labels, so we repurpose similar deconfounding as a diagnostic and find almost no residual signal once style is removed.

\paragraph{Emotion description benchmarks.}
AffectGPT~\cite{lian2025affectgpt} and the MER challenge series~\cite{lian2023mer,lian2024mer,lian2025mer} transitioned emotion recognition from closed-set categorization to open-vocabulary descriptions. EmoPrefer then collected pairwise human preferences over such descriptions, beginning with unanimous three-annotator labels and later scaling to single-annotator pairs~\cite{lian2025emoprefer}; MER2026 adopts the data for its MER-Prefer track alongside six zero-shot multimodal baselines~\cite{lian2026mer,xu2025qwen25omni}. We provide the first systematic analysis of what these preference labels actually encode, and derive concrete recommendations for constructing their future iterations.

\section{Experimental Setup}
\paragraph{Task and data.}
Each instance contains a video $v$, candidate descriptions $(d_1, d_2)$, and a label $\{d_1, d_2, \mathrm{same}\}$. EmoPrefer-V1 has 574 unanimous three-annotator pairs~\cite{lian2025emoprefer}; V2 has 2,096 single-annotator pairs~\cite{lian2026mer}. Candidates are produced by named generators (two in V1; seven in V2), and every V2 pair contrasts distinct generators. Judges predict $d_1$, $d_2$, or ``same'' from the video, audio, and both descriptions, with metadata strictly withheld~\cite{lian2026mer}. Dropping 471 ``same'' pairs, we evaluate the remaining 1,625 (split 1,000/625 by preferred position) via two-class weighted F1 (\waf). All predictors share a five-fold cross-validation split, reporting 2,000-resample bootstrap 95\% confidence intervals (CIs). Content-blind probes read metadata as an oracle; submitted judges infer this solely via writing style. Checkpoint selection by per-fold \waf favors neural judges, making our probe-parity findings conservative.

\paragraph{Models.}
We evaluate two finetuned baselines: (i) a text-only Qwen2.5-7B~\cite{qwen2025qwen25technicalreport} with a LoRA-trained~\cite{hu2022lora} two-class head, and (ii) an audio-visual Qwen2.5-Omni-7B (thinker) with LoRA processing frames, audio, and text. Both use symmetric training and dual-order inference. This is critical: single-order training collapses into severe position bias (30.9 vs.\ 79.7 \waf across opposite orders).

We adapt (iii) an ODIN-style reward model~\cite{chen2024odin} as a diagnostic. A shared LoRA trunk feeds three linear heads summing to the total reward $r(d) = r_C(d) + r_L(d) + r_G(d)$: a content head $r_C$, length head $r_L$, and generator-prior head $r_G$. Using the Bradley--Terry model~\cite{bradley1952rank}, we force $r_C$ to decorrelate from confounds:
\begin{equation}
\begin{aligned}
\mathcal{L} &= \mathcal{L}_{\mathrm{BT}} + \lambda\,\mathcal{L}_{\mathrm{dec}} + \lambda_o\,\mathcal{L}_{\mathrm{orth}},\\
\mathcal{L}_{\mathrm{BT}} &= -\log\sigma\!\big(r(d^{+})-r(d^{-})\big),\\
\mathcal{L}_{\mathrm{dec}} &= |\rho(r_C,\ell)| + |\rho(r_C,\pi)| - \rho(r_L,\ell) - \rho(r_G,\pi),\\
\mathcal{L}_{\mathrm{orth}} &= |\langle w_C,w_L\rangle| + |\langle w_C,w_G\rangle|,
\end{aligned}
\end{equation}
where $\ell$ is length, $\pi$ the generator prior, $\rho$ the batch Pearson correlation, $w_\bullet$ the head weights, and $\lambda = \lambda_o = 1$. $\mathcal{L}_{\mathrm{dec}}$ penalizes $r_C$'s correlation with confounds while forcing $r_L$ and $r_G$ to absorb them; $\mathcal{L}_{\mathrm{orth}}$ maintains orthogonal weights. At inference, confound heads are discarded; $r_C$ scores the pair: $p(d_1\!\succ\!d_2) = \sigma(r_C(d_1) - r_C(d_2))$. We call $r_C$ the \emph{content head}: it captures decorrelated signal but cannot establish video grounding. For audio-visual input, a single-head BT control without decorrelation and the full three-head model both process 12 frames, audio, and one description.

\paragraph{Generator prior and slices.}
For a description $d$ produced by generator $g(d)$, we define the training-fold win-rate prior as:
\begin{equation}
\pi(g) = \frac{\#\{d : g(d)=g,\ d \text{ is preferred}\}}{\#\{d : g(d)=g\}},
\end{equation}
estimated out-of-fold. The prior prediction simply selects the description with the higher $\pi(g)$. We evaluate all predictors across three specific data slices: \textbf{Full}, comprising all non-same pairs; \textbf{Length-matched}, restricted to $\lvert |d_1|-|d_2|\rvert<0.15\max(|d_1|,|d_2|)$ ($n = 310$ in V2) to neutralize verbosity cues; and \textbf{Counter-stereotypical}, pairs where the gold label actively disagrees with the win-rate prior's choice ($n = 552$).

\section{The Audit: What Predicts the Preference?}

\subsection{Content-blind probes match 7B judges}
Table~\ref{tab:main} presents our primary findings. The \emph{length} probe uses character counts, their difference, and log-ratio; the \emph{generator} probe uses one-hot metadata vectors; the \emph{combined} probe uses both.

On Full V2, the combined probe achieves 65.8 [63.2, 68.2] \waf, closely matching the finetuned text judge (66.5) and the audio-visual judge (66.8). The same pattern holds on the length-matched slice (64.4 vs.\ 64.9--65.6). Generator identity heavily outweighs verbosity: identity alone yields 65.5 \waf, whereas length yields 56.1. Because style dominates, evaluating strictly on length-matched data barely dents overall performance.

\subsection{Mechanism: Identity leaks through style, and labels track it}
\label{sec:mechanism}
We identify four key factors driving this performance:
\textbf{(i) Style identifies the generator.} A TF--IDF classifier predicts the source system from text with 99.5\% accuracy (chance 14.3\%).
\textbf{(ii) Identity is always discriminative.} Every pair contrasts different generators; the cue never cancels out.
\textbf{(iii) Labels track an identity prior.} Win rates span 32\% to 79\% and are not explained by length alone (the wordiest system wins only 63\%). A rule tracking the train-fold prior agrees with the gold label on 66\% of pairs, scoring 64.4 on the length-matched slice---matching trained judges.
\textbf{(iv) Identity causally drives the probe.} Randomly shuffling identities collapses the generator probe to the label-blind floor (47.3). Counterfactually swapping test-time identities flips the combined probe's decision on 45\% of pairs, versus 32\% for a length swap (McNemar~\cite{mcnemar1947note} $p < .001$). 

Behavioral analysis confirms judges rely on this cue. On counter-stereotypical pairs, the prior and human label disagree by definition. Trained judges side with the prior on 63--80\% of these pairs, driving the performance down to 20--42 \waf---far below the ${\approx}50$ of random guessing---a systematic bias toward the style prior, not mere noise.

The text ODIN diagnostic (Eq.~1) successfully drives $r_C$'s correlation with length and prior below 0.1 (Fig.~\ref{fig:main}a) and cuts prior-following to a chance-level 48\%. However, the decoupled content head $r_C$ falls to near-chance performance (Table~\ref{tab:main}). Removing decorrelation ($\lambda{=}0$) recovers the 62.4 \waf style shortcut, while chain-of-thought merely interpolates (59\% prior-following) without gaining accuracy. Media access does not change this diagnosis. The AV variant reaches 58.2 \waf without decorrelation (67.6\% prior-following); with decorrelation, prior-following falls to 42.4\% and performance drops to 46.2. Its 58.4 on the counter-stereotypical slice simply mirrors this reversal, leaning against the prior rather than recovering content. (Note: decorrelation might remove legitimate quality signals co-varying with style).

\section{Is There a Recoverable Content Signal?}
\label{sec:negatives}
Can judges exploit audio-visual content when forced? On the length-matched slice, we intervene on judge inputs against the 12-frame+audio baseline (65.3).

\textbf{Media inputs yield no detectable gain.} Text-only and AV judges are indistinguishable (65.6 vs.\ 65.3, McNemar $p = 1.00$). Increasing the audio-visual input from 12 to 16 frames leaves performance flat ($p = 1.00$); even the 8-frame no-audio setup scores 65.0, striking for an emotion benchmark.
\textbf{Reasoning provides no benefit.} A generative-verifier judge~\cite{zhang2025generative}, finetuned on 2,633 STaR rationales~\cite{zelikman2022star}, reaches only 58.9 \waf, below its non-CoT counterpart (65.0).
\textbf{Scaling data does not help.} Adding the V1 split (35\% more data) yields no deconfounded gain (65.2 vs.\ 65.6). V1 is heavily confounded: length- and generator-only probes each reach 79.7 \waf, with only one pair meeting the length-matched criterion.
\textbf{Zero-shot judges inherit the prior.} Zero-shot Qwen2.5-Omni scores 64.3/64.8 (Full/length-matched) and follows the prior 64\% on counter-stereotypical pairs. Without seeing training labels, it clearly shares a stylistic preference with annotators.

Official baselines are strongly bimodal: older models like Qwen2-Audio (36.1) and Video-LLaVA (36.8) sit near the floor, whereas Qwen2.5-VL (76.8) and Qwen2.5-Omni (78.7) reach the high 70s~\cite{lian2026mer}. Checked directly on V2 (Fig.~\ref{fig:main}b), zero-shot Qwen2.5-VL~\cite{bai2025qwen25vltechnicalreport} reproduces the shortcut signature (64.9 Full, 32.8 counter-stereotypical, 67\% prior-following), and LLaVA-NeXT-Video~\cite{zhang2024llavanextvideo} shows similar directional reliance. Prior-following rises with model strength (56\%, 67\%, 73\%): stronger models exploit the source-aligned cue more consistently, while this trend alone does not establish video use.

\section{Recommendations}
The audit exposes three coupled failures: style reveals generator identity, labels track its prior, and little grounded signal survives after deconfounding. The fixes target two stages. At construction, \textbf{source-balanced pairing} preserves cross-generator model comparison but balances winners within each source pairing, while \textbf{strict length control} removes verbosity cues. At evaluation, \textbf{counter-stereotypical and source-pair slices} expose prior-following, while \textbf{multi-annotator consensus} reduces single-rater noise. Thus, generation-time controls weaken the shortcut, and reporting-time controls keep residual source dependence visible. Neither alone establishes that a score is grounded in the video.

\section{Conclusion}
Our audit shows that EmoPrefer's labels are predicted largely by description length and generator identity: a content-blind probe matches finetuned 7B multimodal judges, labels track a per-generator prior, and little scoreable signal remains after ODIN-style deconfounding. EmoPrefer is a valuable step toward open-vocabulary emotion evaluation, but its public formulation conflates \emph{generator-level preference} with \emph{video-grounded accuracy}; agreement does not establish that either description was checked against the video. These conclusions concern the data and judges tested, not what stronger models or re-annotation could recover. As preference modeling guides multimodal alignment, what its labels encode should be audited before their rankings are trusted.

\clearpage
\bibliographystyle{ACM-Reference-Format}
\bibliography{refs_google}

\clearpage
\appendix
\section*{Appendix}
\suppressfloats[t]
This appendix provides the evaluation and implementation details omitted from the main paper. It specifies the fold construction, statistical tests, judge configurations, probe features, and diagnostic controls underlying the results in Table~\ref{tab:main}.

\section{Evaluation Protocol}
\label{app:evaluation}

\paragraph{Data and folds.}
We apply shuffled five-fold \texttt{KFold} (seed 42) to each complete dataset \emph{before} removing ``same'' labels, so all judges and probes share held-out members. V1 contains 574 unanimous three-annotator pairs; V2 contains 2,096 individually labeled pairs. Removing 471 V2 ``same'' labels leaves 1,625 binary pairs. Neither release contains same-generator pairs, and V1 has only one length-matched pair (Table~\ref{tab:taxonomy}).

On V1's 563 binary pairs, five-fold out-of-fold length-only and generator-only logistic probes both score 79.7 \waf (95\% CIs [76.4, 83.1] and [76.3, 83.3]); combining the cues also gives 79.7. Equal aggregate scores do not imply identical predictions. Because V1 has only two generators, every pair is cross-generator, and only one pair is length-matched, the release cannot separate source identity from length through these slices. We therefore use V1 as a confounding check, not as evidence from a deconfounded subset. Adding its 563 pairs enlarges text-judge training by 34.6\%, but evaluation remains on held-out V2 length-matched pairs: combined V1+V2 training scores 65.2 versus 65.6 for V2-only training.

\begin{table}[t]
  \centering
  \caption{Pair taxonomy. Counts except \emph{All} exclude ``same'' labels.}
  \label{tab:taxonomy}
  \begin{tabular}{lrrrrr}
    \toprule
    Data & All & Binary & Same-gen. & Length-m. & Counter \\
    \midrule
    V1 & 574 & 563 & 0 & 1 & 114 \\
    V2 & 2,096 & 1,625 & 0 & 310 & 552 \\
    \bottomrule
  \end{tabular}
\end{table}

\paragraph{Prior and slices.}
For generator $g$, its training-fold win rate $\pi(g)$ is the number of preferred appearances divided by all appearances in binary training pairs. The prior selects the test candidate with larger $\pi(g)$; ties abstain. We recompute $\pi$ from four folds, so test labels never enter the prior or slice definitions. The \emph{length-matched} slice satisfies $\lvert |d_1|-|d_2|\rvert/\max(|d_1|,|d_2|)<0.15$, using character counts; it removes gross verbosity differences, not semantic differences. The \emph{counter-stereotypical} slice contains labels opposing the fold-exclusive prior, so a prior-only rule scores zero there by construction.
On V2, the prior agrees with 1,073 of 1,625 labels (66.0\%), leaving 552 counter-stereotypical pairs. Generator win rates span 32--79\%; the generator with the longest mean output wins 63\%, so source identity is not merely another name for length.

\paragraph{Metric and bootstrap.}
For $c\in\{d_1,d_2\}$, let $F1_c=2P_cR_c/(P_c+R_c)$ and $n_c$ be its support. Following the challenge, we exclude ``same'' labels and report
\begin{equation*}
  \mathrm{WAF}=100\sum_c \frac{n_c}{N}F1_c.
\end{equation*}
This support weighting accommodates the 1,000/625 position-label imbalance; WAF is not accuracy, and its random reference need not equal 50. We concatenate out-of-fold predictions, resample pairs with replacement within each slice, and recompute WAF. The 2.5th/97.5th percentiles of 2,000 resamples (seed 0; one-class resamples skipped) form the CI. It measures finite-pair uncertainty, not training-seed variation; overlapping marginal CIs are not a paired test.

\paragraph{Exact McNemar tests.}
Media configurations predict the same 310 length-matched pairs. Let $b$ count items only judge $A$ gets correct and $c$ those only $B$ gets correct. The two-sided exact test is
\begin{equation*}
  p=\min\!\left\{1,\;2\Pr\!\left[X\leq \min(b,c)\right]\right\},
  \qquad X\sim\mathrm{Binomial}(b+c,0.5).
\end{equation*}
Agreements do not distinguish the judges and therefore do not enter the test. Thus $38/37$ gives $p=1.00$: gains and losses are balanced, providing no detectable improvement at this sample size, not proof of equivalence.

\begin{table}[t]
  \centering
  \caption{Paired tests on the length-matched V2 slice ($n=310$). ``Discordant'' gives the two configurations' exclusive-correct counts.}
  \label{tab:mcnemar}
  \begin{tabular}{lcc}
    \toprule
    Comparison & Discordant & Exact $p$ \\
    \midrule
    Text / 12 frames+audio & 38 / 37 & 1.00 \\
    12 / 16 frames+audio & 14 / 13 & 1.00 \\
    12 frames+audio / 8 frames, no audio & 18 / 17 & 1.00 \\
    \bottomrule
  \end{tabular}
\end{table}

\paragraph{Scope of the media comparisons.}
Text versus 12-frame Omni compares complete judges, not a within-model removal of media. The 12-versus-16-frame row isolates frame budget; the final row changes both frame count and audio. The tests therefore show no detectable gain among these \emph{configurations}, not that media can never help another judge.

\section{Judge and Probe Implementation}
\label{app:implementation}

\paragraph{Shared training choices.}
All finetuned judges use bfloat16 LoRA (rank 16, scale 32, dropout 0.05, learning rate $10^{-4}$) on attention and MLP projections. Text and pairwise Omni checkpoints maximize held-out-fold \waf; ODIN diagnostics use epoch four. This favors neural judges, although it is not nested model selection. Table~\ref{tab:repro} reports the remaining settings; batch entries are microbatch/gradient-accumulation counts.

\begin{table*}[t]
  \centering
  \caption{Training configurations. The Omni single-head control directly optimizes $r_C$ with Bradley--Terry loss; it is not an experiment with fixed auxiliary heads.}
  \label{tab:repro}
  \begin{tabular}{llllccc}
    \toprule
    Configuration & Candidate input & Objective / output & Epochs & Batch/accum. & Context or media \\
    \midrule
    Text judge & $d_1,d_2$ & two-class cross-entropy & 3 & 8/2 & 640 tokens \\
    Omni judge & $v,a,d_1,d_2$ & next-token A/B & 2 & 1/8 & 8, 12, or 16 frames at 1 fps \\
    CoT judge & $v,d_1,d_2$ & rationale then A/B & 2 & 1/8 & 8 frames, no audio \\
    Text, no decorrelation & one $d$ & three-head BT, $\lambda=0,\lambda_o=1$ & 4 & 8/1 & 400 tokens \\
    Text, ODIN & one $d$ & three-head BT, $\lambda=\lambda_o=1$ & 4 & 8/1 & 400 tokens \\
    Omni, no decorrelation & $(v,a,d)$ & single-head BT on $r_C$ & 4 & 2/4 & 12 frames at 1 fps \\
    Omni, ODIN & $(v,a,d)$ & three-head BT, $\lambda=\lambda_o=1$ & 4 & 2/4 & 12 frames at 1 fps \\
    \bottomrule
  \end{tabular}
\end{table*}

\paragraph{Pairwise prompts and order symmetrization.}
Prompts ask which description is more faithful, grounded in what is seen/heard, and human-preferred, returning A/B. Each binary training pair appears in both orders with reversed target. At inference we average $p_A(d_1,d_2)$ and $1-p_A(d_2,d_1)$, mapping both orders to $p(d_1\succ d_2)$. Without symmetric training, the V1 ablation scores 30.9 versus 79.7 \waf in opposite orders, exposing position bias.

\paragraph{Decision extraction.}
The text judge uses a two-class softmax head, whereas Omni compares next-token ``A''/``B'' logits. Both outputs are mapped to $p(d_1\succ d_2)$, averaged across orders, and thresholded at 0.5. ``Same'' items participate only in fold assignment and are masked before binary training and scoring.

Official challenge baseline scores cited in the main text come from the challenge report. The zero-shot rows in Table~\ref{tab:main} are our public-V2 runs with the same dual-order grounding prompt but no EmoPrefer preference finetuning, enabling our diagnostic slices. Their prior-following therefore cannot arise from fitting public preference labels in our run, but does not identify a shared internal mechanism.

Here, a \emph{generative verifier} is the same 8-frame, no-audio Omni backbone trained to generate a rationale before its A/B verdict, rather than predicting A/B directly. In STaR-style rejection sampling, the prompt first requests two or three sentences comparing facial expression, voice, and visible events. We sample up to four rationale--answer sequences per order at temperature 0.7 and retain the first whose final answer matches the training label, yielding 2,633 sequences. The retained rationale and answer become the finetuning target. At test time the model receives no label, generates greedily in both orders, and is scored from its final A/B answer. Thus 58.9 versus 65.0 \waf compares rationale-first and direct-answer training under the same media setting, not different backbone sizes.

\paragraph{Content-blind preference probes.}
These probes predict preference without text or media. For $l_i=|d_i|$, length features are
\begin{equation*}
 [l_1/1000,\;l_2/1000,\;(l_1-l_2)/1000,\;\log((l_1+1)/(l_2+1))].
\end{equation*}
The generator probe concatenates two seven-way one-hot source vectors; adding length gives 18 features. Each is $\ell_2$-regularized logistic regression ($C=1$, 2,000 iterations) fit on four folds. Identity is an audit oracle never given to judges: it measures dataset leakage, not a valid submission.

\paragraph{TF--IDF source classifier.}
This classifier asks whether words reveal \emph{which generator wrote a description}, not which description wins. Each description is a document; its seven-way source is the target. TF--IDF uses at most 20,000 word unigrams/bigrams. Positive term counts become $1+\log \mathrm{count}$; inverse document frequency downweights common phrases; vectors are $\ell_2$ normalized. Logistic regression (2,000 iterations), vocabulary, and IDF are fitted only on training folds. Splitting occurs at the \emph{pair} level before unpacking, so paired descriptions cannot leak across folds. Mean held-out accuracy is $99.5\mathbin{\pm}0.3$\% versus 14.3\% random accuracy. No preference label or media is used: the result shows source recoverability from surface text, not which tokens cause preference.

\paragraph{ODIN-style diagnostic.}
Text encodes one description; Omni encodes video, audio, and one description. Their shared LoRA trunk feeds content, length, and generator heads $(r_C,r_L,r_G)$, whose sum enters Bradley--Terry loss $-\log\sigma(r(d^+)-r(d^-))$. Batch Pearson correlations concatenate both candidates' scores. Decorrelation penalizes $r_C$'s absolute correlation with character length and training-fold prior while assigning those signals to $r_L,r_G$; orthogonality penalizes absolute dot products among head weights. Only $r_C$ is kept at test time. ``Content head'' denotes its intended role, not semantic supervision or proof of grounding.

The text no-decorrelation control retains three heads and orthogonality but sets $(\lambda,\lambda_o)=(0,1)$; Omni instead uses a directly trained single $r_C$. Full diagnostics use $(1,1)$. These controls ask whether decorrelation removes the text shortcut and whether multimodal preference signal exists before decomposition.

\section{Additional Diagnostics}
\label{app:diagnostics}

\paragraph{Permutation and counterfactual controls.}
We jointly permute each pair's $(g_1,g_2)$ relative to labels and refit within every fold. This preserves global source-pair frequencies but breaks identity--preference alignment; across seeds 0--4, generator-only \waf falls from 65.5 to $47.3\mathbin{\pm}0.8$, an empirical label-blind floor rather than theoretical chance. Separately, with the combined probe fixed, swapping only test-time identity vectors flips 45\% of decisions; swapping only length assignments flips 32\%. Paired McNemar comparison of these flip indicators gives $p<.001$. This establishes greater identity control over the \emph{probe}, not how annotators causally formed labels.

\paragraph{Direction of errors.}
``Prior-following'' is the fraction of counter-stereotypical decisions equal to the fold-exclusive prior; the label is opposite by construction. Rates are 79.7\% (text), 73.0\% (12-frame Omni), 69.0\% (16-frame), 73.4\% (8-frame/no-audio), and 59.4\% (CoT), but 48.2\% and 42.4\% for decorrelated text and Omni content heads. The Omni head's higher counter-slice WAF therefore reflects reversal away from the prior, not stable video-grounded recovery.

\end{document}